\documentclass[sigconf]{acmart}

\usepackage{booktabs} % For formal tables
\usepackage{amsmath}
\usepackage{algorithm}
\usepackage[noend]{algpseudocode}
\usepackage{mathtools}
\usepackage{arydshln}
\usepackage{multirow}
\hypersetup{
  colorlinks   = true, %Colours links instead of ugly boxes
  urlcolor     = blue, %Colour for external hyperlinks
  linkcolor    = blue, %Colour of internal links
  citecolor   = black %Colour of citations, could be ``red''
}

\begin{document}

\title{Generalization of Agent Behavior through Explicit Representation of Context}

\author{{\bf Cem Tutum$^1$, Suhaib Abdulquddos$^1$, and Risto Miikkulainen$^{1,2}$}}
\affiliation{$^1$The University of Texas at Austin; $^2$Cognizant Technology
  Solutions\\
  tutum@cs.utexas.edu, suhaib@utexas.edu, risto@cs.utexas.edu\\[2ex]}

%\author{Anonymous Authors}
%\author{Cem Tutum}
%\affiliation{The University of Texas at Austin\\tutum@cs.utexas.edu}
%\author{Suhaib AbdulQuddos}
%\affiliation{The University of Texas at Austin\\suhaib@utexas.edu}
%\author{Risto Miikkulainen}
%\affiliation{Cognizant Technology Solutions;\\The University of Texas at Austin\\risto@cs.utexas.edu}

%\author{Olivier Francon$^{1}$, Santiago Gonzalez$^{1,2}$, Babak Hodjat$^{1}$, Elliot Meyerson$^{1}$,\\
%Risto Miikkulainen$^{1,2}$, Xin Qiu$^{1}$, and Hormoz Shahrzad$^{1}$}
%\affiliation{$^{1}$Cognizant Technology Solutions and $^{2}$The University of Texas at Austin\\[2ex]}

%\renewcommand{\shortauthors}{Tutum, AbdulQuddos, and Miikkulainen}

\begin{abstract}
In order to deploy autonomous agents in digital interactive
environments, they must be able to act robustly in unseen
situations. The standard machine learning approach is to include as
much variation as possible into training these agents. The agents can
then interpolate within their training, but they cannot extrapolate
much beyond it. This paper proposes a principled approach where a
context module is coevolved with a skill module in the game. The
context module recognizes the temporal variation in the game and
modulates the outputs of the skill module so that the action decisions
can be made robustly even in previously unseen situations.  The
approach is evaluated in the Flappy Bird and LunarLander video games,
as well as in the CARLA autonomous driving simulation. The
Context+Skill approach leads to significantly more robust behavior in
environments that require extrapolation beyond training.  Such a
principled generalization ability is essential in deploying autonomous
agents in real-world tasks, and can serve as a foundation for
continual adaptation as well.
\end{abstract}

%\begin{CCSXML}
%<ccs2012>
%<concept>
%<concept_id>10010147.10010257.10010293.10011809.10011812</concept_id>
%<concept_desc>Computing methodologies~Genetic algorithms</concept_desc>
%<concept_significance>500</concept_significance>
%</concept>
%</ccs2012>
%\end{CCSXML}
%
%\ccsdesc[500]{Computing methodologies~Genetic algorithms}
%
%\keywords{Game Playing, Generalization, Extrapolation}

\maketitle

\section{Introduction}

To be effective, autonomous agents need to be able to perform robustly
in previously unseen situations. Especially in real-world decision
making and control applications such as games, simulations, robotics,
health care, and finance, agents routinely encounter situations beyond
their training, and need to adapt safely. A common practice is to
train these models, mostly deep neural networks, with data collected
from a number of hand-designed scenarios. However, the tasks are often
too complex to anticipate every possible scenario, and this approach
is not scalable.

One popular approach to address this problem is few-shot learning, in
particular metalearning, either by utilizing gradients
\cite{Schmidhuber87_PhD,Thrun98_Learn,Finn17_MAML} or evolutionary
procedures \cite{Fernando18_Baldwin,Grbic19_EvoMetaLearn}. In
metalearning, systems are trained by exposing them to a large number
of tasks, and then tested for their ability to learn relevant but
previously unseen tasks. There are also a number of approaches mostly
for supervised learning setting where new labels need to be predicted
based on limited number of training data. However, applications of
few-shot learning to control and decision making, including
reinforcement learning problems, are limited so far
\cite{Kansky17_Schema}.

The approach in this paper is motivated by work on opponent modeling
in poker \cite{Li17_Poker,li:gecco18}. In that domain, an effective
approach was to evolve one neural network to decide what move to make,
and another to modulate those decisions by taking the opponents playing
style into account. When trained with only a small number of very
simple but different opponents, the approach was able to generalize
and play well against a wide array of opponents, include some that
were much better than anything seen during training.

In poker, the opponent can be seen as the context for decision making.
Each decision needs to take into account how the opponent is likely to
respond, and select the right action accordingly. The player can thus
adapt to many different game playing situations immediately, even
those that have not been encountered before. In this paper, this
approach is generalized and applied to control and decision making
more broadly. A skill network reacts to the current situation, and a
context network integrates observations over a longer time period in
parallel. Together they learn to represent a wide variety of
situations in a standardized manner so that a third, controller,
network can make decisions robustly. Such a Context+Skill system can
thus generalize to more situations than any of its components alone.

The Context+Skill approach is evaluated in this paper on several game
domains: (1) Flappy Bird game extended to include more actions and
physical effects (i.e.\ flap forward and drag in addition to flap upward
and gravity); (2) LunarLander-v2 (from OpenAI Gym) extended with
variations in main and side engine power as well as mass of the
lander; and (3) CARLA autonomous driving environment where the
steering curve, torque curve, and map can vary. Such extensions allow
generating a range of unseen scenarios both by extending the range of
effects of those actions as well as their combinations. The approach
generalizes remarkably well to new situations, and does so much better
than its components alone. Context+Skill approach is thus a promising
approach for building robust autonomous agents in real-world domains.

%The remaining sections of this paper are organized as follows: First, the experimental set up and the test domains, the architecture of the neural networks, and the multiobjective evolution procedure for constructing the system are introduced. Next, learning and generalization results are presented, which demonstrates that the Context+Skill approach indeed performs better than its parts. The behaviors of these networks are contrasted in Section of Behavior Analysis, finding that Context+Skill can anticipate results of its actions more accurately, making it possible to adapt to unseen situations. 

%%%%%%%%%%%%%%%%%%%%%%%%%%%%%%%%%%%%%%%%%%%%%%%%%%%%%%%%%%%%%%%%%%%%%%%%%%%%%%%%%

\section{Methodology} \label{sec_method}

This section introduces the experimental domains, the Context+Skill approach, and the evolutionary training methodology.

\subsection{Experimental Domains} \label{Game}

\begin{figure}
\centering
\includegraphics[width=3.25in]{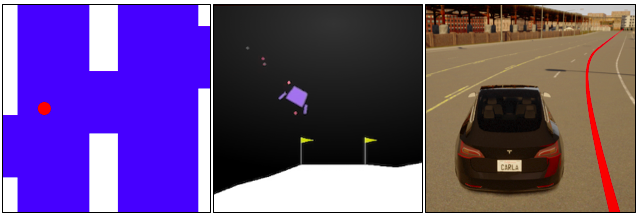}\\
\footnotesize{\hfill($a$) Flappy Ball\hfill($b$) Lunar Lander\hfill\hfill($c$) CARLA\hfill}
\caption{Scenes from Flappy Ball (FB; ($a$)), LunarLander-v2 (LL; ($b$)), and CARLA (($c$). The red circle in the FB domain represents the agent and the white columns are pipes that move from right to left as the game progresses. The purple object in the LL domain represents the spacecraft that controls its main and side engines to land safely on the white surface designated with flags. In CARLA, the agent controls the throttle and steering of a car to reach destination staying as close as possible within the lane indicated by the red curve. The effects of the actions are varied across episodes to evaluate how well the controller adapts to previously unseen situations. For animated examples, see \href{https://drive.google.com/drive/folders/1GBdJzD9tDHJkd59YbQUOIQua6nCiLjXa}{https://drive.google.com/drive/folders/1GBdJzD9tDHJkd59Y\\bQUOIQua6nCiLjXa}.}
\label{fig_domains}
\end{figure}

Three different environments are used for experiments. The first one,
Flappy Ball (FB; Fig.~\ref{fig_domains}$(a)$), is an extension of the
popular Flappy Bird computer game \cite{FlappyBird_wiki} The agent,
controlled by a neural network, aims to navigate through the openings
between pipes without hitting them for a certain length of time. The
agent can flap forward and flap upward; gravity will pull it down and
drag will slow it down. The agent gets a reward of +1 every time it
passes a pipe successfully, and a penalty of -1 at each time step it
crashes into the pipes and -5 when it crashes in the ceiling or
ground. At every time step, the agent receives six-dimensional sensory
information: vertical position,	horizontal and vertical velocities,
horizontal distance to the right edge of the closest pipe, and the
height of the top and bottom pipes, normalized to [0,1]. The
effects of flap upward and forward as well as gravity and drag can
change between episodes; Therefore, the agent has to infer such
variations from its interactions with the environment over time.

The second domain is LunarLander-v2 (LL; Fig.~\ref{fig_domains}$(b)$)
from OpenAI Gym suite, modified to allow variations in mass of the
spacecraft and the effect of the main and two side engine
thrusters. As in the original LL-domain, the agent receives
eight-dimensional numerical sensory information as its input (i.e.,
position and velocity of the lander in 2D, its angle and angular
velocity, and whether each leg touches the ground). The purpose is to
safely land on the lunar surface in a designated region indicated with
the flags. The episode finishes if the lander crashes or comes to
rest.

The third domain is the CARLA open-source autonomous driving
simulation environment (Fig.~\ref{fig_domains}$(c)$). The agent has
both lateral (steering) and longitudinal (throttle) control of the car
while driving between two points in a given certain amount of
time. The steering and the torque curves are modified to evaluate
generalization. Furthermore the agent is placed in completely unseen 
tracks as it tries to complete the same task albeit under more difficult
driving conditions (as imposed by the modified steering and torque curves).
As in the other environments, the agent receives numerical sensory 
information as five numerical values: Rangefinder coordinates which 
describe the distance between the agent and the lane boundaries along five
axes. All control actions are given as continuous values varying within
[-1,1].

%This domain is more complex than the common Flappy Bird game, which does not have the forward flap action or drag. In order to pass more pipes without a collision, the agent needs to use the forward flapping action carefully because the only way to slow down is through drag. It also needs to be cautious because it can only observe the closest pipe. By changing the effects of actions and the forces of gravity and drag, new and more challenging situations can be created, testing the generalization performance of the agent's control policy.

These domains can be seen as proxy for control and decision making problems where the changes in the environment require immediate adaptation, such as operating a vehicle under different weather conditions, configuration changes, wear and tear, or sensor malfunctions. The challenge is to adapt the existing policies to the new conditions immediately without further training, i.e. to generalize the known behavior to unseen situations.

\begin{figure*}
\begin{minipage}{0.3\textwidth}
\centering
\includegraphics[height=1.5in]{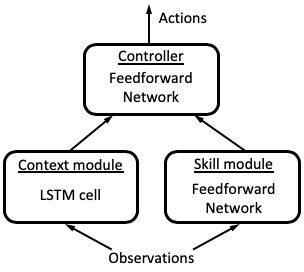}\\
($a$) Context-Skill Network, CS
\end{minipage}
\hfill
\begin{minipage}{0.25\textwidth}
\centering
\includegraphics[height=1.5in]{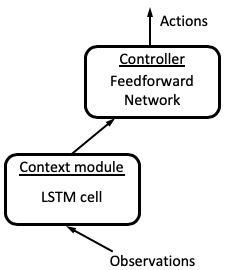}\\
($b$) Context-only Network, C
\end{minipage}
\hfill
\begin{minipage}{0.3\textwidth}
\centering
\includegraphics[height=1.5in]{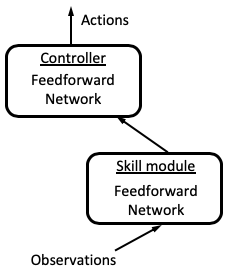}\\
($c$) Skill-only Network, S
\end{minipage}
\caption{The architecture of the Context+Skill network and its ablations. ($a$) The full network consists of three components: a Skill module that processes the current situation, a Context module that integrates observations over the entire task, and a Controller that combines the outputs of both modules, thereby using context to modulate actions. This architecture is compared to ($b$) context-only ablation, and ($c$) skill only ablation in the experiments. Each component is found to play an important role, allowing the CS network to generalize better than its ablations.}
\label{fig_CS}
\end{figure*}

\subsection{Context+Skill Model} \label{sec_net}

The core idea evaluated in this paper is to implement the agent as a Context+Skill network that takes advantage of an explicit representation of context. The Context+Skill Network consists of three components: the Skill and the Context modules and the Controller (Figure~\ref{fig_CS}($a$)). The first two modules receive sensory information from the environment as numerical values, as described in previous subsection. They send their output to the Controller, a fully connected feedforward neural network that makes the decisions on which actions to take.

The Skill module is also a fully connected feedforward network. Together with the Controller they form the Skill-only Network S (Fig.~\ref{fig_CS}($c$)). The Skill module used in this study has 10 hidden and five output nodes and the Controller has 20 hidden hidden nodes. S is used as the baseline model throughout the study. In principle it has all the information for navigating through the pipes, but does not have the benefit of explicit representation of context.

The other main component in the Context-Skill framework is the Context module. It is composed of a vanilla Long Short Term Memory (LSTM) cell (\cite{Hochreiter97_LSTM}). There are three gates in this recurrent memory cell: input, forget, and output. The gates are responsible for learning what to store, what to throw away, and what to read out from the long-term memory of the cell. Thus, the cell can learn to retain information from the past, update it, and output it at an appropriate time, thereby making it possible to learn sequential behavior \cite{Greff17_LSTM,Geron17_MLbook}.

The C-module used in this study consists of an LSTM cell size of 10. The memory of the C-module (h\textsubscript{t-1} and c\textsubscript{t-1}) is reset at the beginning of each new task, and accumulated (transferred) across episodes within each task. It can therefore form a representation of how actions affect the environment. The output of the LSTM (h\textsubscript{t}) is sent to Controller as the context. Together the C-module and the Controller form the Context-only network C (Fig.~\ref{fig_CS}($b$)). It serves as a second baseline, allowing integration of observations over time, but without a specific Skill network to map them directly to action recommendations.

The complete Context-Skill Network, CS (Fig.~\ref{fig_CS}($a$)) consists of both the Context and Skill modules as well as the Controller network of the same size as in C and S. The motivation behind the CS architecture, i.e.\ of integrating the Context module into S, is to make it possible for the system to learn to use an explicit context representation to modulate its actions appropriately. The method for discovering these behaviors is discussed next.

\subsection{Evolutionary Learning Process}

The goal in each domain is to find a safe solution that optimizes the performance objective. Although it is possible to formulate the optimization process based on that single objective, it turns out the diversity resulting from the multiobjective search speeds up training and helps discover well-performing solutions \cite{Knowles01_LocalOpt}.  In the experiments in this paper, the solution that is perfectly safe or close to it is returned as the final result, and its generalization ability evaluated.

In each domain, the first objective, $f_0$, measures safety, and the second, $f_1$, measures performance, and is in conflict with safety.
Accordingly in the FB domain, the number of any type of collisions (or hits) is minimized, whereas the number of successfully passed pipes is maximized. In the LL-domain, the total rewards (indicating a successful landing) is maximized, whereas the landing time is minimized. In the CARLA domain, safety $f_0$ is optimized by minimizing a cumulative weighted sum of a lane-distance penalty and a wobbliness penalty:
\begin{equation}
f_0=\sum_{t=1}^{t=T} (d(t) + \lambda \mathrm{abs}(s(t) - s(t-1))),
\end{equation}
where $d(t)$ is the distance from the center of the lane at time $t$ (shown as a red line in Fig.~\ref{fig_domains}($c$)), $s(t)$ is the steering output at time $t$, and $\lambda$ is a proportionality coefficient ($=5.5$ in the experiments below). Time is incremented in 1/30 sec intervals; each episode lasts $T=20$ seconds, or less if the agent reaches a specified target zone before that.
Performance $f_1$ is optimized by minimizing the Euclidean distance between the agent and the target zone at the end of the episode.

\begin{table}
\vspace*{-1ex}
\caption{Parameter ranges during training}
\centering
\label{table_baseVals}
\begin{tabular}{|c|c|c|}
\hline
FB ($\pm$20\%)  & LL ($\pm$10\%) & CARLA ($\pm$15\%)                              \\ \hline
Flap\textsubscript{base} = -12.0  & Main\textsubscript{base} = 20.0 & $\alpha$\textsubscript{base}=1.0 \\
Fwd\textsubscript{base} = 5.0     & Side\textsubscript{base}=1.0  & $\beta$\textsubscript{base}=1.0  \\
Gravity\textsubscript{base} = 1.0 & Mass\textsubscript{base}=8.0  &                                     \\
Drag\textsubscript{base} = 1.0    &                     &        \\       \hline     
\end{tabular}
\end{table}

In the evolutionary learning process, the weights of all three neural networks described in Section~\ref{sec_net} are evolved while the network architecture remains fixed. The goal is to maximize the average fitness across multiple tasks, where each task is based on different physical parameters in the FB, LL and CARLA domains. The FB-domain has four parameters (Flap, Fwd, Gravity, Drag), the LL-domain has three parameters (Main, Side, Mass) and the CARLA-domain has two parameters ($\alpha$ and $\beta$, which control the steering angle and torque curves of the car, respectively). In each task during evolution, only one parameter is subject to change, while the rest are fixed at their base values (given in Table~\ref{table_baseVals}). Thus, the number of tasks is equal to the number of parameters. The parameters in each domain are varied during training within $\pm$20\%, $\pm$10\% and $\pm15$\% in the FB, LL and CARLA-domains, respectively.

Each task, and therefore each parameter, is uniformly sampled n\textsubscript{episodes}=5 (except in the CARLA domain where each parameter is sampled six times) times within the limits specified above. Thus, there are 20 fitness evaluations per individual in each generation in the FB domain, 15 evaluations in the LL-domain and 12 evaluations in the CARLA-domain. The fitness of every individual in the population, i.e., average score of all episodes, is evaluated in parallel on the same task distribution for a fair comparison. The memory of C-module in CS and C is reset at the beginning of each task, and transferred from episode to episode otherwise.

\begin{algorithm}[t]
\caption{Evolutionary process for training networks}\label{algo_evo}
\begin{algorithmic}[1]
\Procedure{evolve}{}
   \State \scalebox{0.9}{stop := False}
   \State \scalebox{0.9}{parents := random\_init\_individuals($\mu$)}
   \State \scalebox{0.9}{task\_params = prepare\_task\_params(n$_{\mathrm{episodes}}$,n$_{\mathrm{tasks}}$)}
   \State \scalebox{0.9}{ fitness = distribute(eval\_fitness(), (parents, task\_params)) }
   \For{gen \textbf{from} 1 \textbf{to} n$_{\mathrm{gen}}$}
   		\State \scalebox{0.9}{offspring = tournament\_sel\_DCD(parents, $\mu$)}
       \For{i \textbf{from} 1 \textbf{to} $\lambda$}
             \If {random() $\leq$ p$_{\mathrm{crossover}}$}
                 \State \scalebox{0.9}{SBX(offspring[i], offspring[i+1])}
             \EndIf
             \State \scalebox{0.9}{Polynomial\_Mutation(offspring[i])}
             \State \scalebox{0.9}{Polynomial\_Mutation(offspring[i+1])}
       \EndFor
       \State \scalebox{0.9}{params = prepare\_task\_params(n$_{\mathrm{episodes}}$,n$_{\mathrm{tasks}}$)}
       \State \scalebox{0.85}{ fitness = distribute(eval\_fitness(), (parents, task\_params)) }
       \For{j \textbf{from} 1 \textbf{to} $\lambda$}
            \If {fitness[j][0] $\geq$ pipes$_{\mathrm{max}}$}
                \If {fitness[j][1] $\leq$ hits$_{\mathrm{max}}$}
                     \State \scalebox{0.9}{stop := True}
                \EndIf
            \EndIf
       \EndFor
       \State \scalebox{0.9}{ parents := tournament\_sel\_DCD(parents + offspring, $\mu$) }
       \If {stop == True}
            \State \Return parents
            \State \textbf{break}
       \EndIf
   \EndFor
\EndProcedure
\end{algorithmic}
\end{algorithm}

Non-dominated sorting genetic algorithm (NSGA-II) \cite{Deb02_NSGA2} was implemented in DEAP \cite{DEAP} as the optimization method. The overall procedure is shown in Algorithm~\ref{algo_evo}. It receives n\textsubscript{tasks}, n\textsubscript{episodes}=5, perturb=0.2 in FB, 0.1 (in LL) or n\textsubscript{episodes}=12, perturb=0.15 (in CARLA), base parameter values, $\mu$ = 96 (48 in CARLA), p\textsubscript{crossover} = 0.9, n\textsubscript{gen} = 2,500 as input. The population size ($\mu$) is chosen as a multiple of 24 since the fitness evaluations are distributed among 24 threads on a cluster (i.e., Dell PowerEdge M710, 2$\times$ Xeon X5675, six cores @ 3.06GHz). Default NSGA-II settings, including SBX (Simulated Binary Crossover), Polynomial Mutation, tournament\_sel\_DCD (Tournament Selection Based on Dominance), and ($\mu$ + $\lambda$) elitist selection strategy were used \cite{Deb02_NSGA2}. The process is robust to minor variations of these choices.

%\begin{figure}[ht]
%\centering
%\includegraphics[width=3.25in]{figures/fig_CS_vs_S_carla}\\
%\mbox{~}\hfill$f_0$ (min deviation)\hfill\hfill $f_1$ (min distance)\hfill\mbox{~}\\
%CS - S
%\vspace*{-2ex}
%\caption{Comparing generalization of Context+Skill and its Skill-only ablation in the CARLA domain. Both networks %reach the same acceptable level of safety (i.e.\ deviation), but CS generalizes much better along the performance %objective (i.e.\ distance).}
%\label{fig_carla_results}
%\end{figure}

%\begin{figure}
%\includegraphics[width=3.25in]{figures/fig_boxplot}
%(a) $f_0$ (pipes)\hspace{0.15\textwidth} (b) $f_1$ (hits)\\
%\caption{Summary of the generalization distributions. The data from Fig.~\ref{histograms} is organized into boxplot so that the distributions for the different architectures can be compared more clearly. CS generalizes better than both C and S, which are rather similar. CS thus combines the abilities of both C and S for superior generalization.}
%\label{fig_boxplot}
%\end{figure}

%%%%%%%%%%%%%%%%%%%%%%%%%%%%%%%%%%%%%%%%%%%%%%%%%%%%%%%%%%%%%%%%%%%%%%%%%%%%%%%%%%
\section{Results} \label{sec_results}

The evolutionary learning results are first described, followed	by
evaluation of the generalization ability of the	best solutions.
For animated examples of behaviors in each domain, see \href{https://drive.google.com/drive/folders/1GBdJzD9tDHJkd59YbQUOIQua6nCiLjXa}{https://drive.google.\\com/drive/folders/1GBdJzD9tDHJkd59YbQUOIQua6nCiLjXa}.

\subsection{Learning} \label{sec_learning}

CS, C, and S architectures were evolved with different random seeds
six times in FB, five times in LL, and once in CARLA (due to the computational complexity of this domain).
To avoid overfitting, specific stopping criteria
were selected for each domain after some experimentation. In FB,
training was run until an individual in the population achieved a
fitness scores of at least $f_0$=0.01 (hits) and $f_1$=22.0 (pipes).
In LL, training was run until an individual reached $f_0$=200
(total rewards).  In CARLA, evolution was run until an individual
reached $f_0$=2043 (safety penalty) and $f_1$=55 meters (distance from the target). In FB and CARLA a minimum performance criteria was included to make sure the agent would not simply optimize safety by staying still. 
The final Pareto-optimal set in each run
contained multiple individuals; in FB and LL a safe network (i.e.\
satisfying the stopping criteria) with the highest performance, and in
CARLA the network with the least Euclidean distance to the origin in the Pareto front, was selected as the final network, and its generalization ability evaluated.

The evolution of S in general took the shortest amount of generations
since it has the least number of parameters (287 compared with 982 in
C and 1207 in CS). The same architecture is used for all three
domains. To make sure the number of parameters was not a factor,
another S with a larger Skill module, with the same number of
parameters as CS, was also evolved with	the same stopping criterion.
However, it performed poorly compared to the smaller S in the
generalization studies, apparently because it was easier to
overfit. Thus, it was excluded from the comparisons.

\subsection{Generalization} \label{sec_generalization}

To evaluate the generalization performance of the best performing networks, the task parameters in each domain were changed in the following ways:

\begin{itemize}
\item The range of variation in the task parameters was increased significantly beyond the training limits (Table~\ref{table_generalization});
\item All parameters were varied simultaneously instead of one at a time; and
\item In the CARLA domain, the agents were evaluated on an entirely new track that had never been encountered during training. This track included curves with significantly higher curvature than those encountered during training.
\end{itemize}

\begin{table}[t]
%\vspace*{-2.5ex}
\caption{Parameter ranges during generalization}
\centering
\label{table_generalization}
\begin{tabular}{|c|c|c|}
\hline
FB ($\pm$75\%)  & LL ($\pm$50\%)  & CARLA ($\pm$35\%) \\ \hline
Flap\textsubscript{min} = -21.0  & Main\textsubscript{min} = 10.0 & $\alpha$\textsubscript{min} = 0.65 \\
Flap\textsubscript{max} = -3.0  & Main\textsubscript{max} = 30.0 & $\alpha$\textsubscript{max} = 1.35 \\ \hdashline
Fwd\textsubscript{min} = 1.25     & Side\textsubscript{min} = 0.5  & $\beta$\textsubscript{min} = 0.65  \\
Fwd\textsubscript{max} = 8.75     & Side\textsubscript{max} = 1.5  & $\beta$\textsubscript{max} = 1.35  \\ \hdashline
Gravity\textsubscript{min} = 0.25 & Mass\textsubscript{min} = 4.0  &                                     \\
Gravity\textsubscript{max} = 1.75 & Mass\textsubscript{max} = 12.0  &                                     \\ \hdashline
Drag\textsubscript{base} = 0.25    &                     &        \\
Drag\textsubscript{base} = 1.75    &                     &        \\       \hline
\end{tabular}
\end{table}

\begin{figure*}
\begin{minipage}{0.3\textwidth}
\centering
\includegraphics[width=\textwidth]{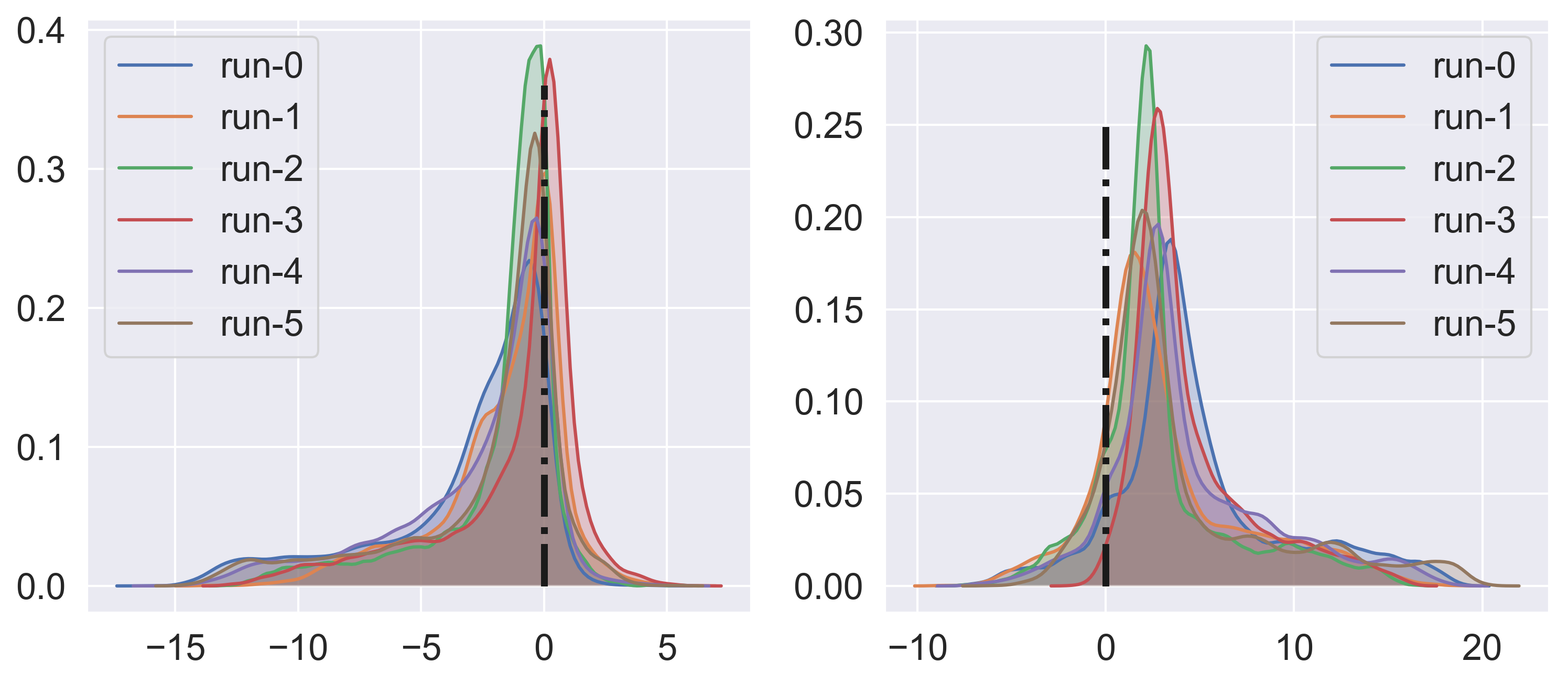}
$f_0$ (min hits)\hspace{0.2\textwidth} $f_1$ (max pipes)\\
($a$) CS - S
\end{minipage}
\hfill
\begin{minipage}{0.3\textwidth}
\centering
\includegraphics[width=\textwidth]{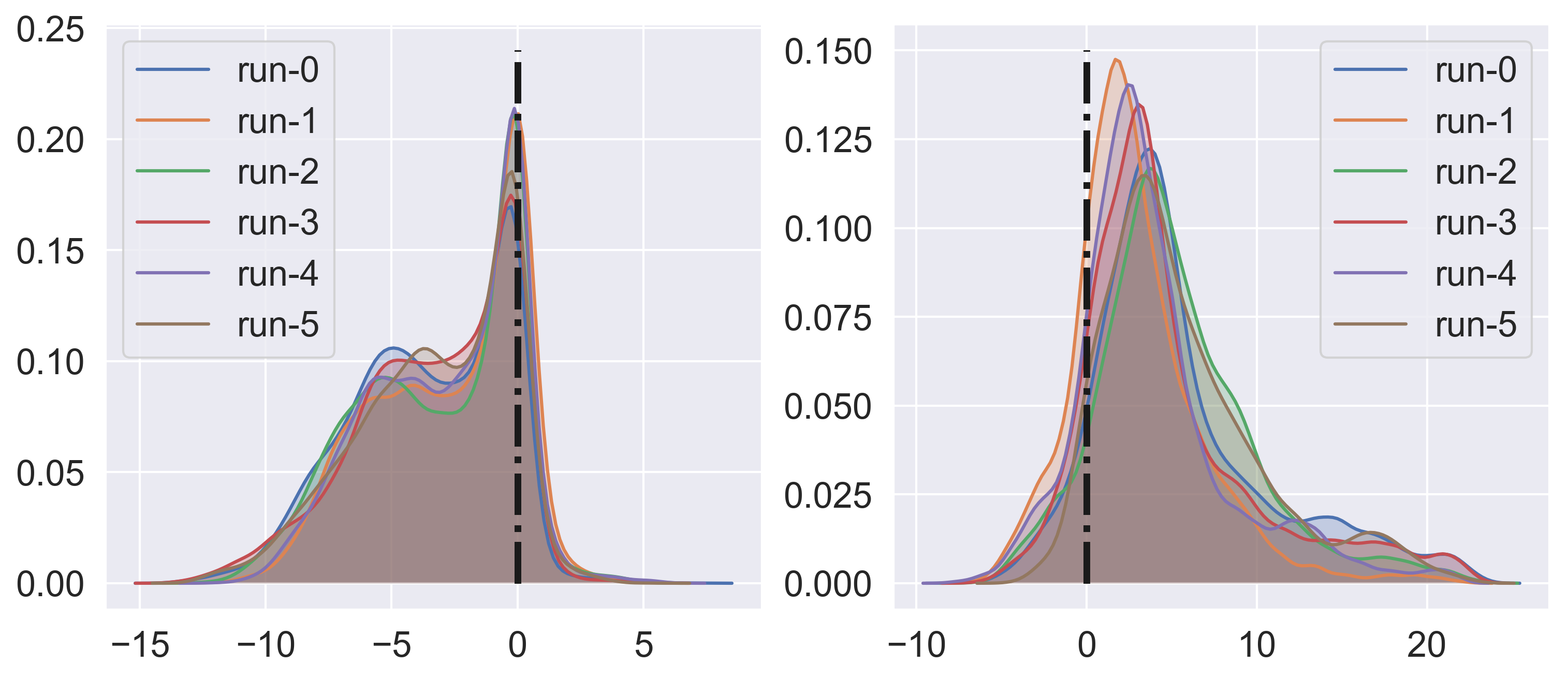}
$f_0$ (min hits)\hspace{0.2\textwidth} $f_1$ (max pipes)\\
($b$) CS - C
\end{minipage}
\hfill
\begin{minipage}{0.3\textwidth}
\centering
\includegraphics[width=\textwidth]{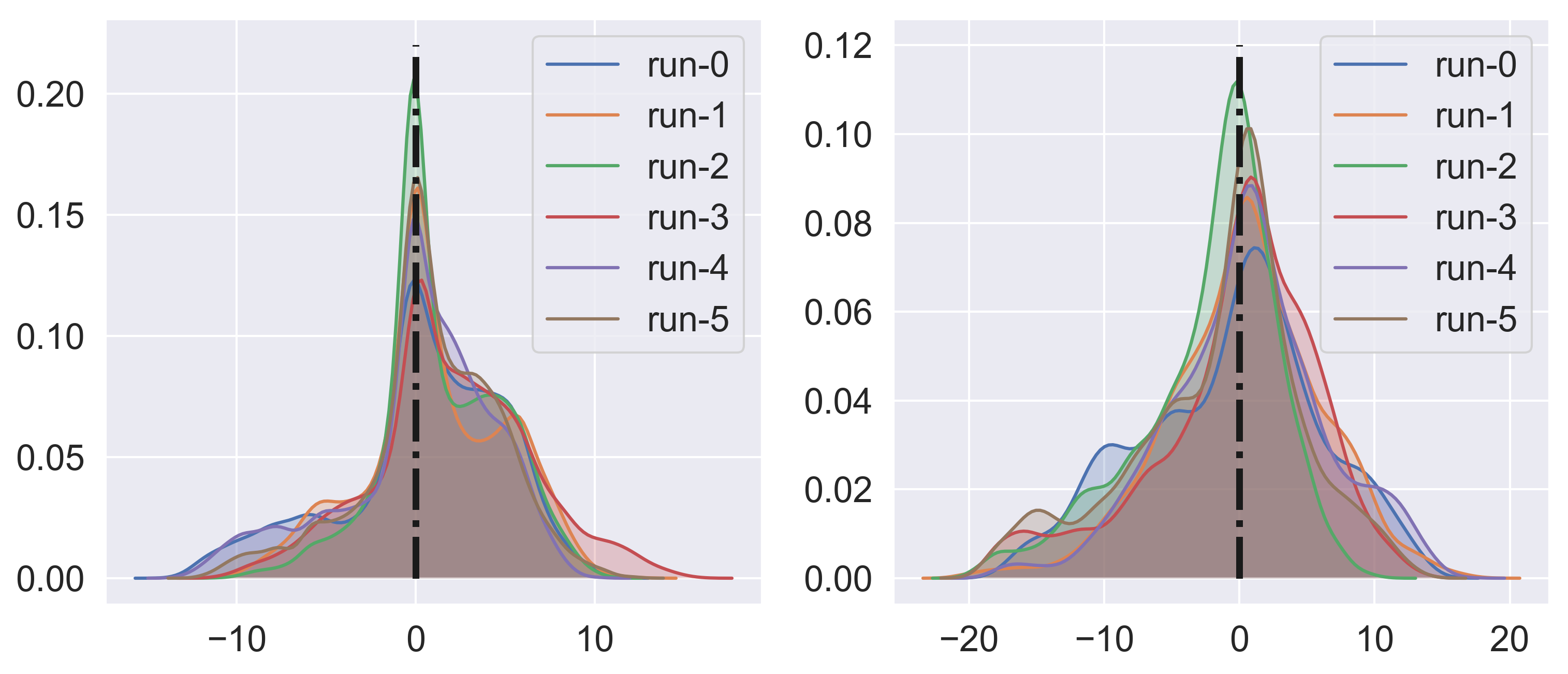}
$f_0$ (min hits)\hspace{0.2\textwidth} $f_1$ (max pipes)\\
($c$) C - S
\end{minipage}
\vspace*{-2ex}
\caption{Comparing generalization of Context+Skill and its ablations in the FlappyBall domain. The $x$-axis shows the differences in generalization performance across the $3\times10^4$ test episodes for the three pairs of architectures. With minimization objectives, a distribution that is skewed to the left of the 0 line (black dashed line) is better, and with maximization, a distribution that is skewed to the right is better. CS generalizes much better than S ($a$) and C ($b$), which are about equal ($c$).}
\label{histograms1}
\vspace*{2ex}
\end{figure*}

\begin{figure*}
\begin{minipage}{0.3\textwidth}
\centering
\includegraphics[width=\textwidth]{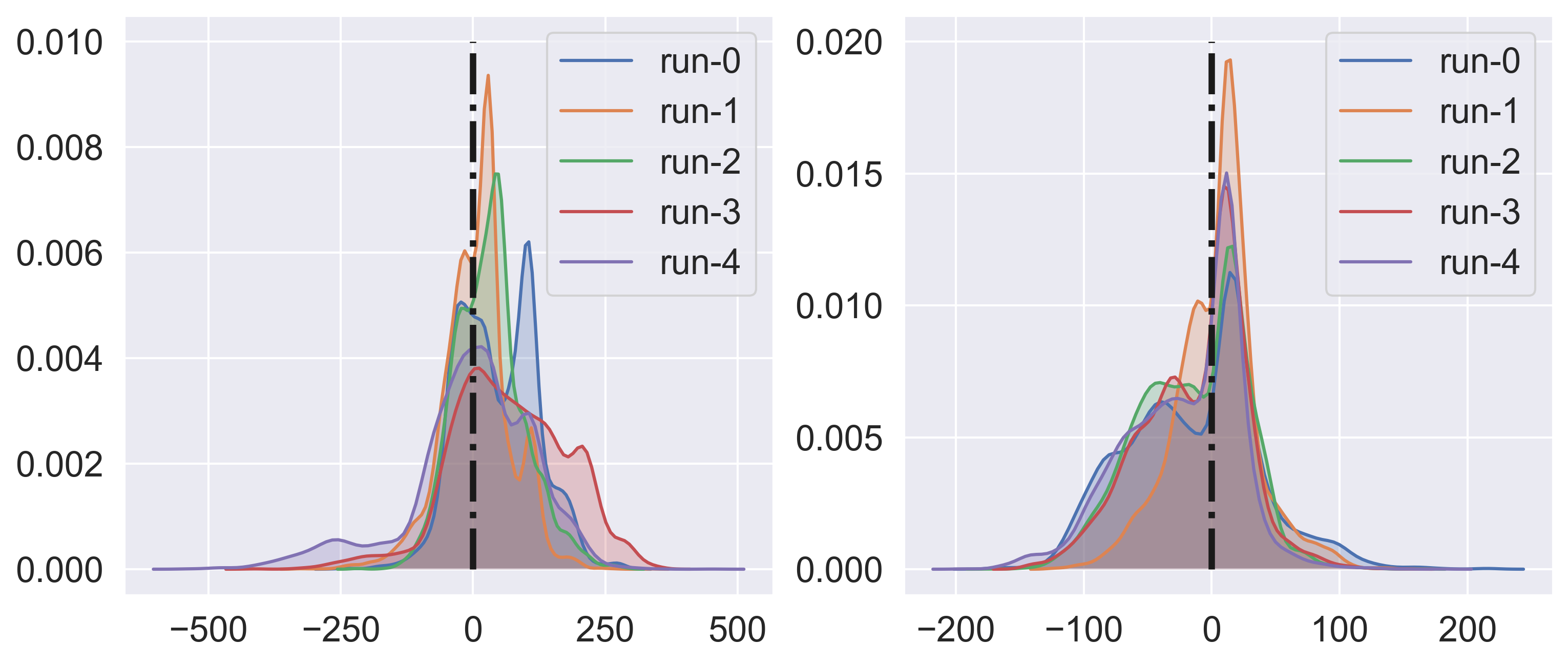}
$f_0$ (max rewards)\hspace{0.15\textwidth} $f_1$ (min time)\\
($a$) CS - S
\end{minipage}
\hfill
\begin{minipage}{0.3\textwidth}
\centering
\includegraphics[width=\textwidth]{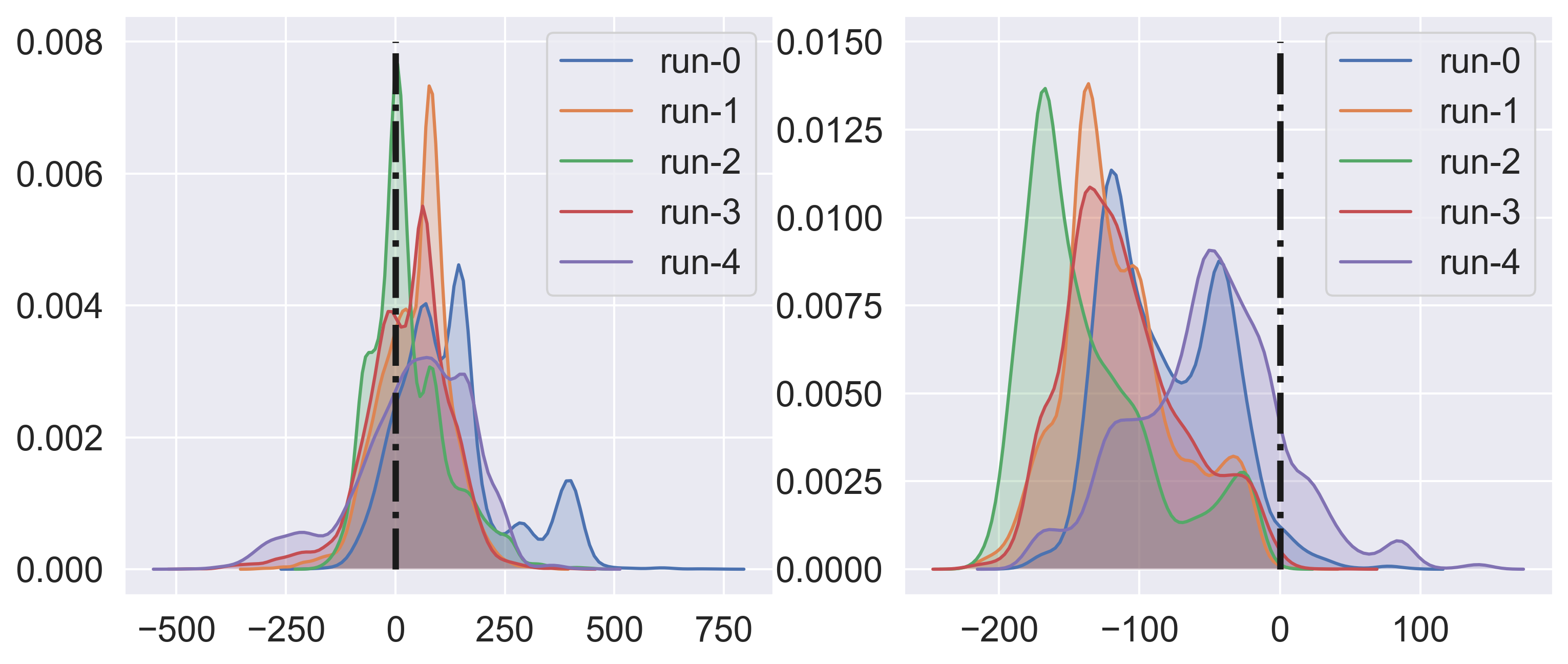}
$f_0$ (max rewards)\hspace{0.15\textwidth} $f_1$ (min time)\\
($b$) CS - C
\end{minipage}
\hfill
\begin{minipage}{0.3\textwidth}
\centering
\includegraphics[width=\textwidth]{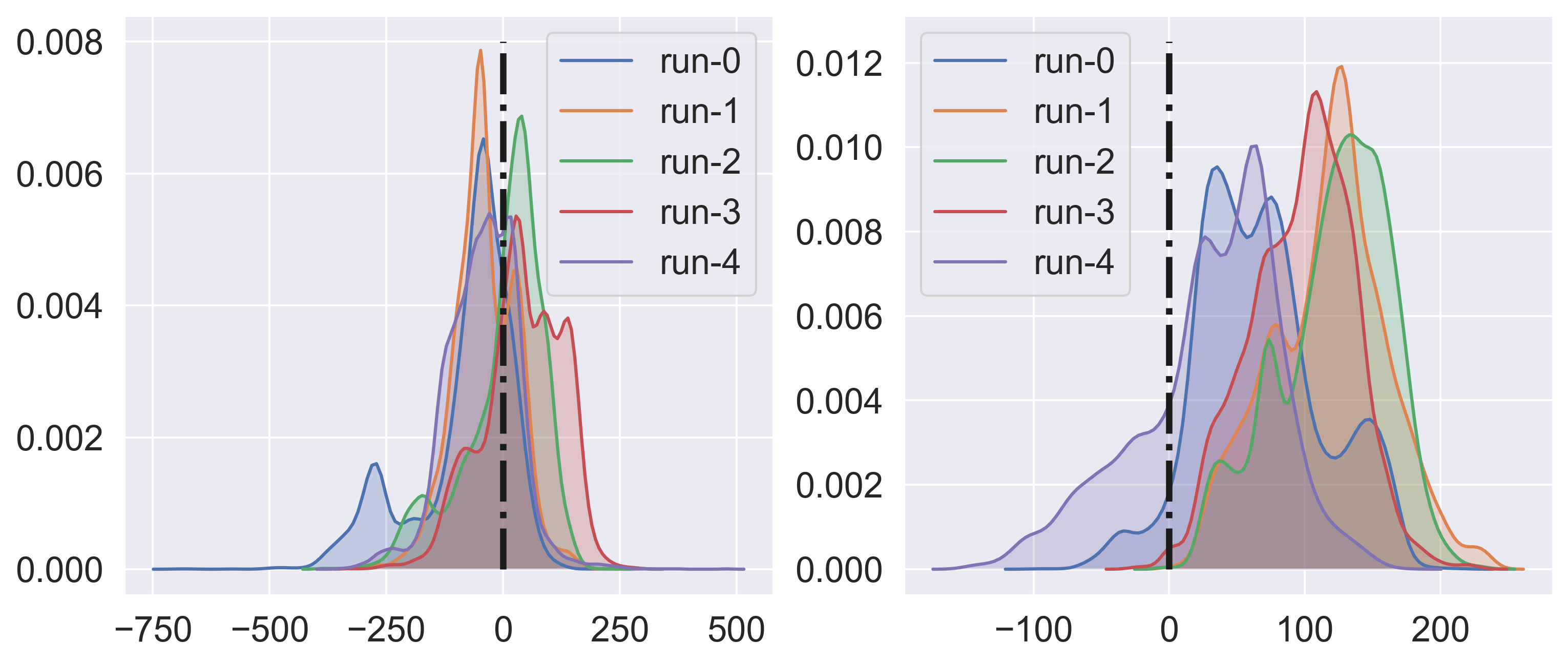}
$f_0$ (max rewards)\hspace{0.15\textwidth} $f_1$ (min time)\\
($c$) C - S
\end{minipage}
\vspace*{-2ex}
\caption{Comparing generalization of Context+Skill and its ablations in the LunarLander domain. As in FlappyBall, CS generalizes much better than S ($a$) and C ($b$), which are about equal ($c$).}
\label{histograms2}
\vspace*{2ex}
\end{figure*}

\begin{figure*}
\begin{minipage}{0.32\textwidth}
\centering
\includegraphics[width=\textwidth]{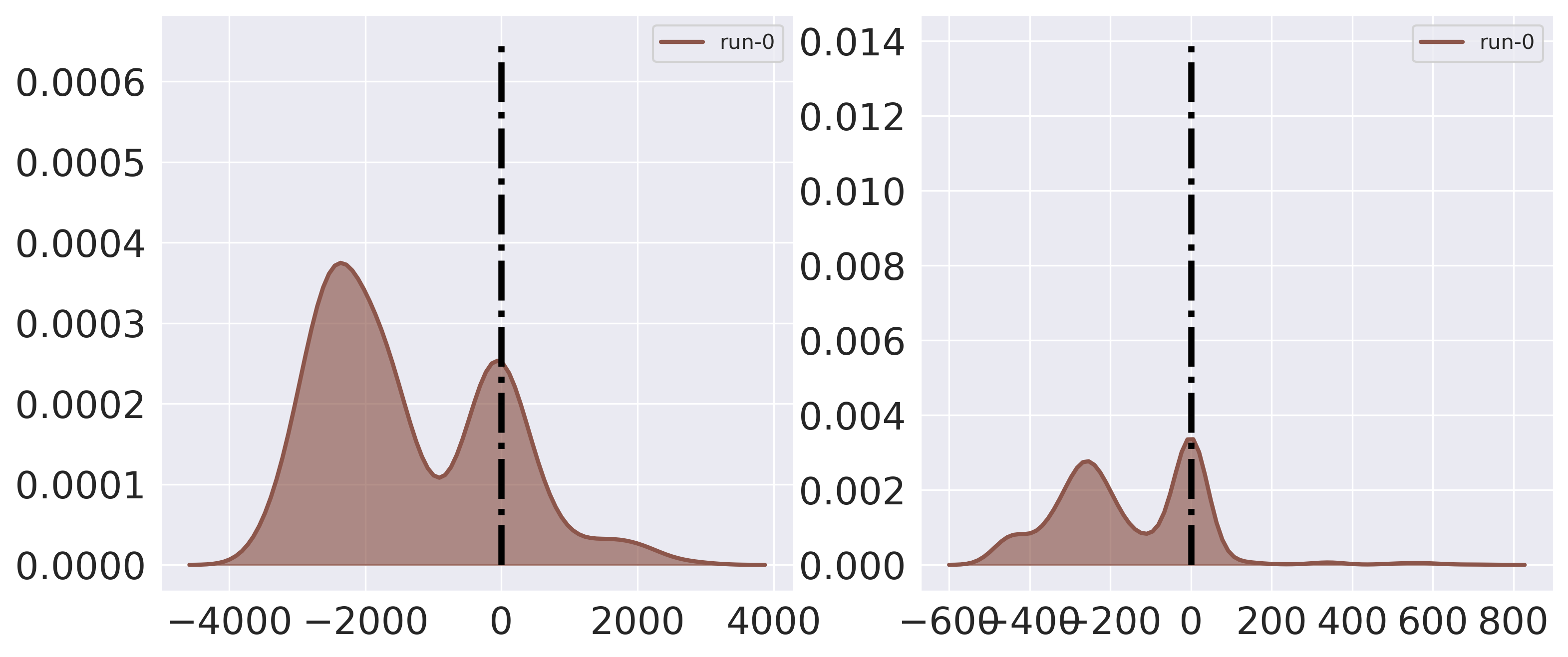}
$f_0$ (min safety)\hspace{0.15\textwidth} $f_1$ (min distance)\\
($a$) CS - S
\end{minipage}
\hfill
\begin{minipage}{0.32\textwidth}
\centering
\includegraphics[width=\textwidth]{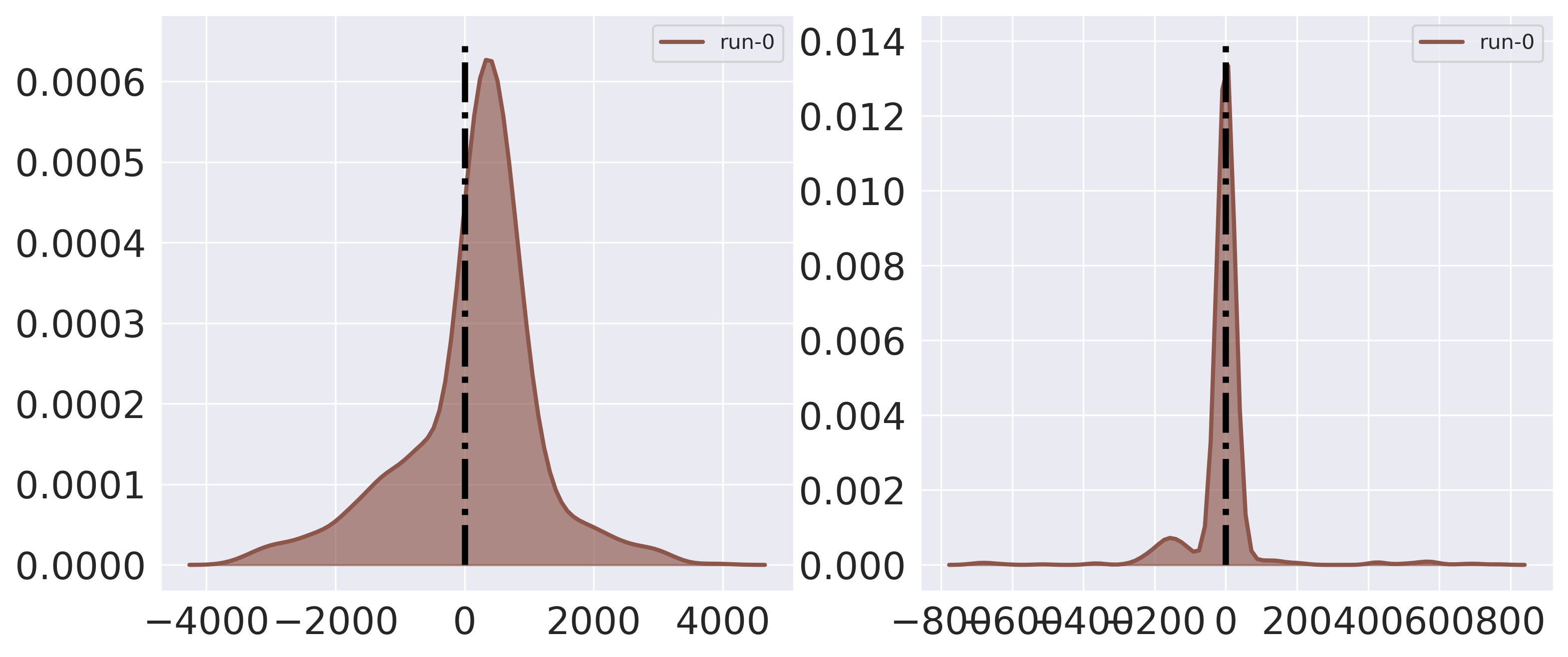}
$f_0$ (min safety)\hspace{0.15\textwidth} $f_1$ (min distance)\\
($b$) CS - C
\end{minipage}
\hfill
\begin{minipage}{0.32\textwidth}
\centering
\includegraphics[width=\textwidth]{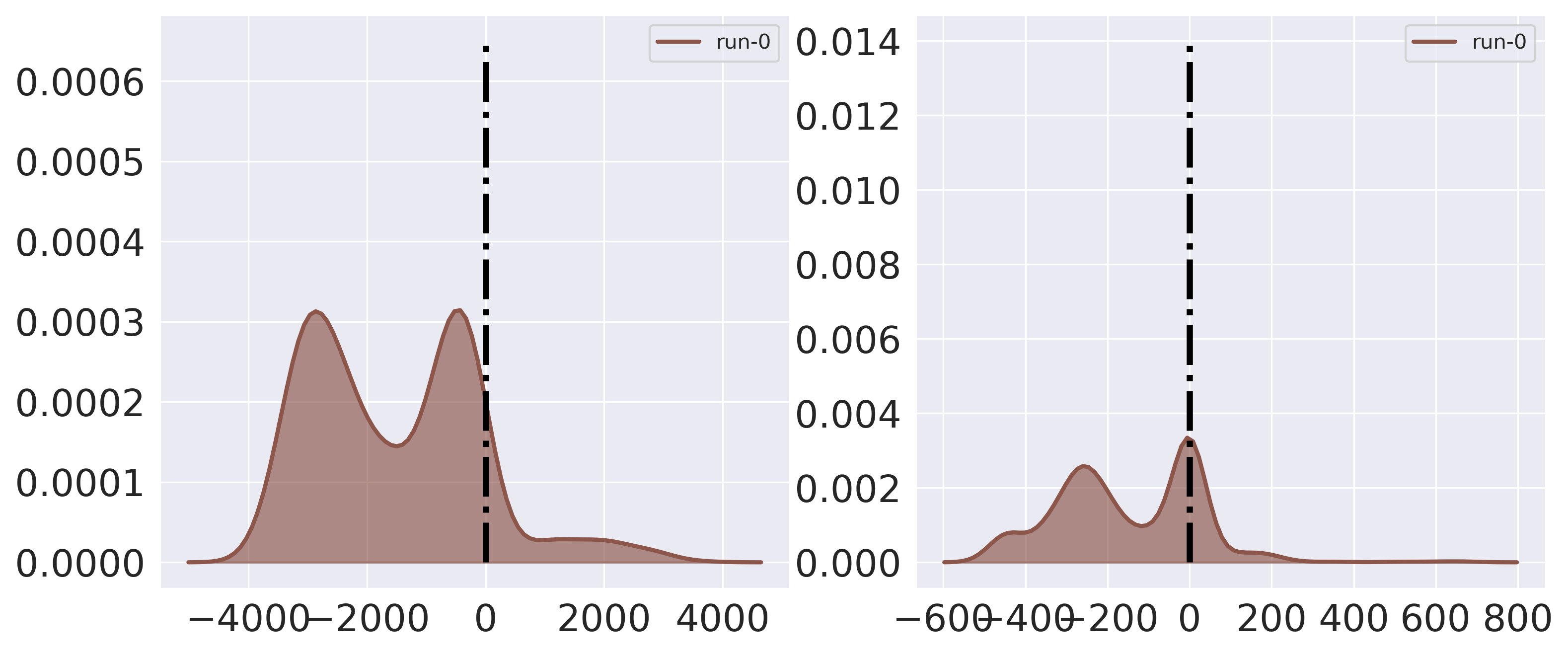}
$f_0$ (min safety)\hspace{0.15\textwidth} $f_1$ (min distance)\\
($c$) C - S
\end{minipage}
\vspace*{-2ex}
\caption{Comparing generalization of Context+Skill and its ablations in the CARLA domain. As in the other two domains, CS generalizes much better than S ($a$) and is slightly better in performance and about equal in safety than C ($b$). C in this domain performs better than S ($c$), suggesting that context dynamics are more important than reactivity in this domain.}
\label{histograms3}
\vspace*{2ex}
\end{figure*}

More specifically, each parameter axis was divided into equal steps
(i.e.\ 10, 20 and 35 in FB, LL and CARLA, respectively). In FB and LL, each set
of task parameters were sampled three times and averaged; in CARLA, only one sample was used. Therefore,
all three networks were tested for $3\times10^4$, $24\times10^3$ and
$1.225\times10^3$ episodes in these domains.
Figures~\ref{histograms1},~\ref{histograms2},
and~\ref{histograms3} show the difference in generalization
between CS and S, CS and C and C and S. They are histograms indicating
how often each difference was observed in the episodes.	The first
network in the subtraction performs better wrt.\ a minimization
objective if the histogram is skewed to the left, and wrt.\ a
maximization objective if the histogram is skewed to the right.

Thus, the plots show that in both FB and LL, CS generalizes much better than S and C both in terms of safety and performance. In CARLA, CS generalizes better than S and C in terms of performance; in terms of safety, CS generalizes much better than S in terms of lane distance and about the same in wobbliness, and it generalizes much better than C in terms of wobbliness and about the same or slightly less in terms of lane distance. Thus, CS consistently achieves superior performance and mostly superior safety across multiple domains.

\begin{figure*}[ht]
\centering
\includegraphics[width=5in]{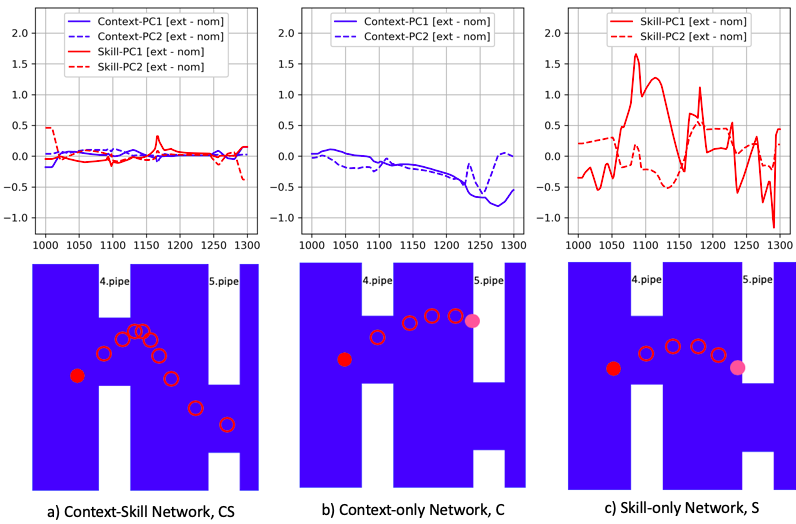}
\caption{Contrasting the generalization ability of ($a$) CS, ($b$) C, and ($c$) S networks. Bottom row: At the fourth pipe in the generalization task, S and C flap upward and then forward, end up too high too fast without enough time to come back down, crashing into the fifth pipe. In contrast, the Context-Skill Network avoids the collision by correctly estimating the effects of its actions, giving itself enough time to come down. Top row: Differences between the principal components of the module outputs between the nominal and generalization tasks. They differ little in CS, making it easier for the controller to output the correct actions in new contexts. For an animation of these episodes, see \href{https://drive.google.com/drive/folders/1GBdJzD9tDHJkd59YbQUOIQua6nCiLjXa}{https://drive.google.com/drive/folders/1GBdJzD9tDHJkd59YbQUOIQua6nCiLjXa}.}
\label{fig_behavior}
\end{figure*}

%%%%%%%%%%%%%%%%%%%%%%%%%%%%%%%%%%%%%%%%%%%%%%%%%%%%%%%%%%%%%%%%%%%%%%%%%%%%%%%%%%
\section{Behavior Analysis}
\label{behavior}

To understand how the CS architecture outperforms its individual components C and S, an FB task with parameters [Flap=-7.0, Gravity=0.58, Fwd=8.75, Drag=0.58] was evaluated further. This setting has a previously unseen exaggerated effect for forward flap, and a previously unseen diminished effects for upward flap, gravity and drag. Thus, actions tend to push up and speed up the agents more than expected, and it is difficult for it to slow down and come down.

Neither the C nor the S network performed well in this task: The C network collided with six pipes and S with five. Remarkably, CS managed to pass all 21 pipes. Both C and S used all four actions (flap upward, forward, simultaneously upward and forward, and glide, i.e.\ do nothing), but CS interestingly never uses flap upward alone. That action simply lifts the agent up, which is rarely optimal action in this environment where it takes such a long time to come down. If it is necessary to go up it is because the opening is high, and in that case it is more efficient to move forward at the same time.

As an illustration, second row in Fig.~\ref{fig_behavior} shows a situation at the fourth and fifth pipe. Both C and S make a similar mistake by flapping up and forward. They end up too high too fast, do not have enough time to come back down, and crash into the fifth pipe. In contrast, even before the fifth pipe becomes visible, CS refrains from both actions while there is enough time for weaker gravity and drag to slow and pull down the agent, and it reaches the opening in the fifth pipe just fine.

To understand how CS manages to implement this behavior whereas C and S do not, the outputs of the C and S modules are compared between this generalization task (where C and S hit the fifth pipe) and a task where all parameters are at their base values (where C and S do not hit the fifth pipe).  Their 10 and 5-dimensional outputs are first reduced to two dimensions through principal component analysis and then subtracted. The top row of Fig.~\ref{fig_behavior} shows these differences for the C- and S-modules of CS, for the C-module of C, and for the S-module of S, at the locations in the image below.

One might intuitively expect that C-module in C and S-module in S would not change their behavior much in the generalization task, but the C-module in CS would vary significantly to modulate the output of the S-module. Surprisingly, the opposite is true: Both the C and the S-module in CS vary very little compared to those in C and S (see also the quantitative comparison in Table~\ref{table_behavior}). The generalization task presents novel inputs that results in novel outputs in C and S, and the controller does not know how to map them to correct actions. In contrast, the C and S-modules in CS have learned to standardize their output despite the change in context; their outputs are what the controller expects as its input, and is able to output the correct actions. Interestingly, this effect is similar to standardizing context in sentence processing, which makes it possible to generalize to novel sentence structures \cite{miikkulainen:spec}. Remarkably, whereas in sentence processing the standardization was implemented by a hand-designed architecture, in Context+Skill it is automatically discovered by evolution.

\begin{table}
\centering
\caption{Change in C and S modules during generalization.}
\label{table_behavior}
\begin{tabular}{|c|c|c|c|}
\hline
Network                  & PC          & MSD   & STD   \\ \hline
                         & Context-PC1 & 0.004 & $\pm$0.058 \\
\multirow{2}{1em}{CS} & Context-PC2 & 0.003 & $\pm$0.039 \\
                         & Skill-PC1   & 0.007 & $\pm$0.082 \\
                         & Skill-PC2   & 0.018 & $\pm$0.133 \\ \hline
\multirow{2}{1em}{C}  & Context-PC1 & 0.139 & $\pm$0.282 \\
                         & Context-PC2 & 0.061 & $\pm$0.147 \\ \hline
\multirow{2}{1em}{S}  & Skill-PC1   & 0.414 & $\pm$0.600 \\
                         & Skill-PC2   & 0.088 & $\pm$0.286 \\ \hline
\end{tabular}
\end{table}

\section{Discussion and Future Work}

The Context+Skill approach represents context explicitly, and has a
remarkable ability to generalize to unseen situations.  In the
experiments so far, the neural networks have a fixed topology; it may
be possible to customize their architecture further through evolution
\cite{Stanley02_NEAT,Schrum14_Modular}, and thereby delineate and
optimize their roles further.  Besides the architecture, the choice of
training tasks plays an important role;	methods that automatically
design a curriculum, i.e., a sequence of new training tasks
\cite{Narvekar18_CL,Wang19_POET,Schmidhuber11_Powerplay,Justesen18_AutoCL,Risi19_PCG},
could lead to further improvements. Third, instead of using
handcrafted features, convolutional layers added in front of the
Context and Skill modules could be used to discover features while
training, extending the approach to more general visual tasks.

Lifelong machine learning tries to mimic how humans and animals learn
by accumulating the knowledge gained from past experience and using it
to adapt to new situations incrementally
\cite{Parisi18_Continual}. The generalization ability of Context+Skill
can serve as a foundation for continual learning. It provides an
initial rapid adaptation to new situations upon which further learning
can be based. How to convert generalization into a permanent ability
in this manner is an interesting direction of future research.

\section{Conclusion}

A major challenge in deploying artificial agents in the real world is
that they are brittle---they can only perform well in situations for
which they were trained. This paper demonstrates a potential solution
based on separating contexts from the actual skills. Context can then
be used to modulate the actions in a systematic manner, significantly
extending the unseen situations that can be handled. This principle
was successfully evaluated in three domains: challenging versions of
the Flappy Bird and LunarLander-v2 games as well as the CARLA
autonomous driving simulation. The results suggest that the
Context+Skill approach should be useful in many control and decision
making tasks in the real world.

\section{Acknowledgments}
This research was supported in part by DARPA L2M Award DBI-0939454.

\small
\bibliographystyle{ACM-Reference-Format}
% \bibliography{main}


\end{document}